\documentclass[journal,twoside]{IEEEtran}
\usepackage{cite}
\usepackage{amsmath,amssymb,amsfonts}
\usepackage{algorithm}

\usepackage{algpseudocode}
\usepackage{graphicx}
\usepackage{textcomp}
\usepackage{booktabs}

\usepackage{framed,multirow}
\usepackage{minted}

\usepackage{enumitem}   % safe now
\usepackage{url}
\usepackage{xcolor}

\usepackage{amsmath}
\usepackage{hyperref}
\usepackage{cleveref}
\crefname{algorithm}{algorithm}{algorithms}
\usepackage{subcaption}
\definecolor{newcolor}{rgb}{.8,.349,.1}

\def\BibTeX{{\rm B\kern-.05em{\sc i\kern-.025em b}\kern-.08em
    T\kern-.1667em\lower.7ex\hbox{E}\kern-.125emX}}

\begin{document}

\title{Random Window Augmentations for Deep Learning Robustness in CT and Liver Tumor Segmentation}

\author{Eirik A. Østmo, Kristoffer K. Wickstrøm, Keyur Radiya, Michael C. Kampffmeyer,\\ Karl Øyvind Mikalsen, and Robert Jenssen
\thanks{Submitted October 2025. This work was supported by the Research Council of Norway [grants 309439, 315029, 303514], the Pioneer Centre for AI, DNRF [grant P1], and Sigma2 with compute [grant NN8106K].}
\thanks{E. A. Østmo, K. K. Wickstrøm, M. C. Kampffmeyer, K.Ø. Mikalsen, and R. Jenssen are with the UiT Machine Learning Group, Tromsø, Norway (e-mail: \{eirik.a.ostmo, kristoffer.k.wickstrom, michael.c.kampffmeyer, karl.o.mikalsen, robert.jenssen\}@uit.no).
K. Radiya and K. Ø. Mikalsen are with the University Hospital of North Norway, Tromsø, Norway (e-mail: keyur.radiya@unn.no).
M. C. Kampffmeyer and R. Jenssen are also with the Norwegian Computing Centre, Oslo, Norway, and R. Jenssen is also with Pioneer Centre for AI, Copenhagen, Denmark.}
}

\maketitle

\begin{abstract}
Contrast-enhanced Computed Tomography (CT) is important for diagnosis and treatment planning for various medical conditions. Deep learning (DL) based segmentation models may enable automated medical image analysis for detecting and delineating tumors in CT images, thereby reducing clinicians' workload. Achieving generalization capabilities in limited data domains, such as radiology, requires modern DL models to be trained with image augmentation. However, naively applying augmentation methods developed for natural images to CT scans often disregards the nature of the CT modality, where the intensities measure Hounsfield Units (HU) and have important physical meaning. This paper challenges the use of such intensity augmentations for CT imaging and shows that they may lead to artifacts and poor generalization. To mitigate this, we propose a CT-specific augmentation technique, called Random windowing, that exploits the available HU distribution of intensities in CT images. Random windowing encourages robustness to contrast-enhancement and significantly increases model performance on challenging images with poor contrast or timing. We perform ablations and analysis of our method on multiple datasets, and compare to, and outperform, state-of-the-art alternatives, while focusing on the challenge of liver tumor segmentation.

\end{abstract}


%% main text
\section{Introduction}

% Importance of CT in medical imaging and liver tumor applications
Computed Tomography (CT) is a cornerstone in the diagnosis and treatment planning of various health conditions \cite{rubinComputedTomographyRevolutionizing2014}. In liver applications, contrast-enhanced CT imaging enables precise imaging for detection and delineation of tumors, facilitating effective intervention strategies.

% Deep learning
With the rapid advancement of Deep Learning (DL), the utilization of computer vision (CV) models has become increasingly prevalent for automating tasks in radiology \cite{wangComprehensiveSurveyDeep2024, hansenADNetFewshotLearning2023, estevaDeepLearningenabledMedical2021, litjensSurveyDeepLearning2017}. 

% Need for CT liver, vessel, and tumor segmentation
With novel techniques and improved accuracy of recent DL based segmentation models, the potential for impactful clinical applications emerges.
Limited data has been a longstanding challenge in DL \cite{chingOpportunitiesObstaclesDeep2018} and liver tumor applications \cite{radiyaPerformanceClinicalApplicability2023}, and techniques such as image augmentation have proven to be indispensable in enhancing the generalization capabilities of CV models \cite{shortenSurveyImageData2019}. Intensity augmentations stochastically distort the pixel distribution of an image and thus prevent overfitting in DL models. However, the widespread extension of such image augmentation across modalities, and specifically the use of augmentation schemes developed for natural images in the field of CT and medical imaging, raises concerns.

% Motivation
In the field of medical imaging, modalities like CT have absolute pixel intensities that convey physical meaning, and conserving them is important in DL applications \cite{isenseeNnUNetSelfconfiguringMethod2021}. Consequently, we hypothesize that naively applying intensity augmentations may prevent the model from learning intensity-related features and hurt performance.

As the medical imaging community increasingly relies on DL methods for tasks such as segmentation \cite{jeonFullyautomatedMultiorganSegmentation2024, liuTumorConspicuityEnhancementbased2024, zhouAutomaticSegmentationMultiple2020}, classification \cite{manjunathDeepLearningAlgorithm2024, riquelmeDeepLearningLung2020}, and disease detection \cite{abdulahiPulmoNetNovelDeep2024, liDeepLearningApplications2022} in CT images, the need for robust and domain-specific augmentation strategies becomes paramount. Failing to acknowledge and adapt DL methods to the characteristics of CT scans may compromise the efficacy and reliability of DL models, potentially leading to erroneous patient diagnoses and treatment.

\begin{figure}
    \centering
    \includegraphics{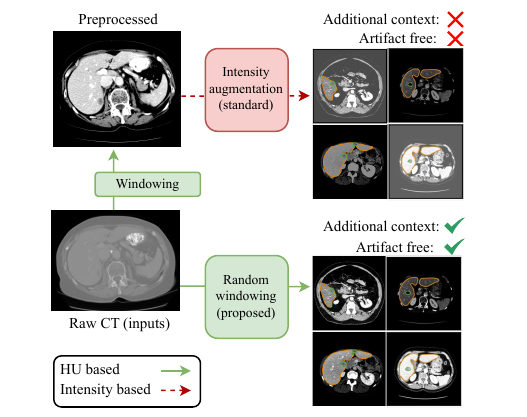}
    \caption{Standard intensity augmentation of CT images often operates on the clipped intensities of the image. This limits the augmentation potential and available context and may create artifacts in the image, like unnatural values for background, bone, or air pockets. We propose Random window augmentations for CT that operate on the raw HU using the viewing window, which resolves the aforementioned challenges.}
    \label{fig:graphical_abstract}
\end{figure}

% Objective
To address these challenges, this paper introduces novel CT-specific augmentation techniques (\autoref{fig:graphical_abstract}) to replace and improve upon widely adopted intensity augmentations. In this pursuit, this paper challenges the widely adopted use of intensity augmentations developed for natural images in DL applications for CT imaging \cite{isenseeNnUNetSelfconfiguringMethod2021, hatamizadehUNETRTransformers3D2022}. Scrutinizing the effect of CT preprocessing and popular intensity augmentations leads to a novel CT augmentation technique, \textit{Random windowing}, that outperforms and can replace preceding methods in contrast-enhanced CT images and liver tumor applications.

\subsection{Contributions}
We summarize the main contributions of this paper:
\begin{itemize}
    \item We introduce \textit{Random windowing},
    a CT-specific augmentation scheme that encourages robustness and can be targeted to specific regions.
    \item We thoroughly analyze and ablate the effects of Random windowing, its components, and alternatives on contrast-enhanced CT images for liver tumor segmentation.
    \item Random windowing is compared to state-of-the-art alternatives and is found to yield models with stronger performance on challenging CT images that suffer from poor intravenous contrast or poor contrast timing.
\end{itemize}

\subsection{Outline}
In Section \ref{sec:related_work}, we present related work that our methods complement and build upon. 
Section \ref{sec:methodology} introduces Random windowing, with its effects analyzed in Section \ref{sec:analysis_of_random_windowing}. Results and ablations that validate our method are presented in Section \ref{sec:results}, followed by discussion and a future outlook in Section \ref{sec:discussion}.

\section{Related work}
\label{sec:related_work}
\subsection{Preprocessing of CT images}
In a CT image, the measured volumetric linear attenuation $\mu$ of scattered X-rays are calibrated against the attenuation of water $\mu_{\text{water}}$ and air $\mu_{\text{air}}$, resulting in intensity units measured in Hounsfield units (HU) given by
\begin{equation}
    HU = 1000 \cdot \dfrac{\mu - \mu_{\text{water}}}{\mu_{\text{air}} - \mu_{\text{water}}}.
    \label{eq:hounsfield units}
\end{equation}
Before CT images are visualized, they are often preprocessed to a \textit{viewing window}, by clipping the intensities to a given range, resulting in increased contrast of the region of interest (ROI). Although DL models can take unprocessed HU as inputs \cite{vorontsovLiverLesionSegmentation2018}, they often benefit from clipping the intensity values to a narrower range \cite{bilicLiverTumorSegmentation2023, isenseeNnUNetSelfconfiguringMethod2021, hanAutomaticLiverLesion2017}. The benefit comes from increased relative HU differences within the ROI at the cost of removing certain intensities assumed to be irrelevant.

For CT in general, and liver tumor segmentation specifically, there is much variation in the chosen clipping range, which may suggest that a suboptimal window is common \cite{bilicLiverTumorSegmentation2023}. The clipping boundaries in DL applications are often determined from radiology domain knowledge \cite{islamEvaluationPreprocessingTechniques2021}, computed from intensity statistics of the dataset \cite{isenseeNnUNetSelfconfiguringMethod2021}, or determined dynamically during training \cite{zakrzewskiAdaptiveHounsfieldScale2024}. In our experiments, we show that choosing a narrow, task-specific clipping range is beneficial for segmentation performance.

\begin{figure*}
    \centering
    \includegraphics[width=\linewidth]{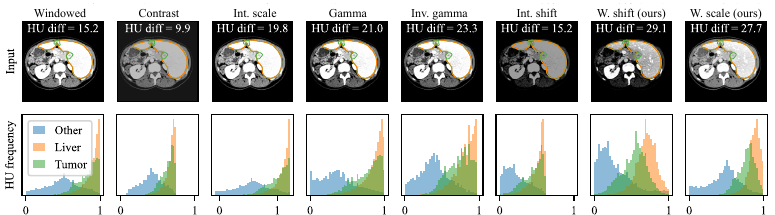}
    \caption{On certain contrast-enhanced CT images, standard preprocessing removes important information about liver and tumor intensities. Standard image transformation applied to such preprocessed images fails to reintroduce useful variation into the image. Our proposed windowing augmentations are applied before any preprocessing and have the potential to yield better visualizations of such difficult images.}
    \label{fig:individual_augs}
\end{figure*}

In contrast-enhanced CT, contrast injected into an upper extremity vein highlights abdominal tissues, with the arterial phase (20–30 s) showing liver arteries and the portal venous phase (50–70 s) enhancing liver parenchyma by $\geq 50$ HU \cite{shibamotoInfluenceContrastMaterials2007}. Due to the sensitive timing of contrast-enhancement, the variation in ROI appearance and HU of the same phase can be great across patients and scans. DL-based CT applications often rely on image augmentation to learn robustness to these variations.

\subsection{Augmenting CT images}
Data augmentation involves applying various transformations to existing training data to create slightly altered instances of the data, which enrich the dataset to enhance the model's robustness and generalization \cite{shortenSurveyImageData2019}. For medical images, two main types of augmentations are especially relevant: geometric augmentations and intensity augmentations. 

\textit{Geometric augmentations} preserve the pixel intensities by only altering the spatial appearance using geometric transformations like rotation, flipping, translation, resizing, and cropping. \textit{Intensity augmentations} transform the pixel values of the image without changing the spatial aspects of the image. 
Certain augmentations, such as saturation and hue transformation, operate in the RGB space of natural images and require three color channels, making them unsuitable for CT images, which have HU in only one channel (grayscale). Intensity augmentations like contrast, brightness, and gamma corrections, however, can be applied to CT intensity values to change the visual appearance of the image.

Geometric augmentations are commonly used in DL applications for CT images \cite{chlapReviewMedicalImage2021} as well as in liver and tumor applications \cite{bilicLiverTumorSegmentation2023}. Applying geometric augmentations like flip, rotation, translation, crop and resize, for CT can accommodate for lack in variation of orientation, shape, and sizes of tumors and other anatomical structures. Patch-based training inherently provides translation variability by exposing the model to structures at different spatial positions, while also enabling computational memory benefits \cite{isenseeNnUNetSelfconfiguringMethod2021}.

Intensity augmentations for DL in CT applications are not always required for good performance, as many well-performing methods manage fine without them \cite{hanAutomaticLiverLesion2017, vorontsovDeepLearningAutomated2019, bilicLiverTumorSegmentation2023}. However, many top-performing methods leverage some forms of intensity augmentations \cite{bassiTouchstoneBenchmarkAre2025,
liuCLIPDrivenUniversalModel2023a,
antonelliMedicalSegmentationDecathlon2022,
tangSelfSupervisedPreTrainingSwin2022, isenseeNnUNetSelfconfiguringMethod2021, chenTransUNetRethinkingUNet2024} to increase variability in limited data domains. The most popular intensity augmentations are intensity shifting and scaling methods, closely connected to contrast and brightness augmentations for natural images. 

\subsection{Questionable augmentation practices}
Shifting and scaling raw CT intensity values is not problematic in a DL setting, but could simulate variations in measurements that could naturally occur across scans, protocols, and patients. We argue that the problem arises when such intensity augmentations are applied to \textit{clipped} intensity values.

When HU are clipped to a viewing window, relevant for the application, the information outside the viewing window is removed and is not possible to recover. Subsequent scaling and shifting during brightness and contrast transformations will risk introducing artifacts, in the form of empty values, near the edges of the interval, instead of simulating natural variation (\autoref{fig:individual_augs}).

While we acknowledge that many CT applications might already apply intensity augmentations with care, we consider the importance of this to be understated.
The nnU-Net \cite{isenseeNnUNetSelfconfiguringMethod2021} augmentation pipeline leverages a combination of brightness, contrast, and gamma augmentation from Batchgenerators \cite{isenseeBatchgeneratorsPythonFramework2020}, and has been reused in multiple CT applications \cite{chenTransUNetRethinkingUNet2024, huang3DCrossPseudoSupervision2022, wubbelsDeepLearningAutomated2025}. The Unetr \cite{hatamizadehUNETRTransformers3D2022} and Swin-Unetr \cite{tangSelfSupervisedPreTrainingSwin2022} apply intensity shifting and scaling from the MONAI framework \cite{cardosoMONAIOpensourceFramework2022}. These top-performing segmentation frameworks all apply intensity augmentation \textit{after} HU clipping, which we find concerning.

Although these augmentations seemingly increase performance, we hypothesize that augmentation strategies that are tailored towards CT and treat the HU distribution of CT with care are more advantageous.

\subsection{CT-specific augmentations}
CT-specific intensity augmentations have in common that they leverage the HU distribution more cautiously. Cluster-wise voxel intensity range shift \cite{kloenneDomainspecificCuesImprove2020} applies additional predefined region-specific or manufacturer-specific viewing windows after initial windowing to further focus the model on specific parts of the CT distribution. Similar strategies that sample between predefined and task-specific viewing windows have been proposed independently as augmentation strategy or as a training technique multiple times since: \cite{tanDataAugmentationCNN2021a} 
%\citet
generated samples from three predefined viewing windows and used them to train a COVID classifier from lung CT images. Similarly, \cite{limTransRAUNetDeepNeural2025} 
% \citet
used images preprocessed with four predefined viewing windows as augmentation for liver vessel segmentation, and found it to be favorable over geometric transformation such as flipping and mirroring.
\cite{abboudImpactTrainTesttime2023} investigated the effect of using various predefined viewing windows in training and inference in liver lesion segmentation and found that window selection is important for segmentation performance.
Tangential to augmentation, works exploiting multiple inputs with images of different viewing windows during training have also been explored in segmentation \cite{kwonTrainableMulticontrastWindowing2020} and self-supervised learning \cite{wickstromClinicallyMotivatedSelfsupervised2023}.

While these methods avoid artifacts, they do not provide the continuous properties comparable to traditional augmentation techniques. They also do not address the issue of patient, contrast, or timing variations introduced by the contrast-enhancement in diagnostic CT scans.

We propose to \textit{continuously} vary the viewing window used for preprocessing by sampling the window width and level randomly. The augmentation strength can be tailored for the relevant task by controlling the allowed range of viewing windows. Our method, entitled \textit{Random windowing}, creates training images that can simulate difficult cases and make difficult cases easier for the model, resulting in increased robustness. Contrary to traditional intensity augmentations applied to preprocessed HU values, our method does not introduce artifacts from shifting and scaling pre-clipped HU.

We show that our proposed approach for CT augmentation improves robustness and generalization across multiple architectures and datasets, and for liver tumor segmentation in both the 2D and 3D settings. Additionally, we show that some of the traditional augmentation schemes, not respecting the unit of intensity in the CT modality, in fact can hurt performance in certain settings.

\section{Methodology: Window augmentations for CT images}
\label{sec:methodology}
In this section, we introduce our new CT augmentation technique, Random windowing, as well as the core components of the technique. Specifically, the windowing operation used for preprocessing, Window shifting, and Window scaling. These operations together make up our CT augmentation method, Random windowing.

\subsection{Windowing operation}
\textit{Windowing} is a preprocessing scheme for CT images and is an essential step performed by radiologists upon CT inspection and in CT DL applications. It removes irrelevant information by limiting the range of HU to display.

Windowing narrows down the HU distribution by clipping the values to a minimum and maximum value. The viewing window is defined by the window width $W$ and the window level $L$. The width $W$ determines how much of the HU range to include, and the level $L$ is the center of the range. For each application or task, a base viewing window, comprising a base width $W_\text{base}$ and base level $L_\text{base}$, is typically selected to optimize visualization. The included HU intensities $x$ are then given by:

The included HU intensities $x$ in the preprocessed image are given by
\begin{equation}
    x \in [L - \frac{W}{2}, L + \frac{W}{2}].
    \label{eq:windowing}
\end{equation}

After windowing, the range of intensity values to display is smaller, and thus fewer values are mapped to each grayscale level in the display. The contrast of the image is therefore increased, so details are more prominent to both radiologists and DL models (\autoref{fig:graphical_abstract} "Windowing").
For liver tumor segmentation, we find, in Section \ref{sec:effect_of_base_viewing_window}, that a narrow tumor-specific window is beneficial for performance.

\subsection{Window shifting}
\label{sec:window_shifting}
When a narrow viewing window is selected, the CT images are more affected by varying contrast-enhancement from timing of the IV contrast and the patient's response to it.

To mitigate this problem, Window shifting\footnote{Window shifting was first introduced in the conference version of this paper \cite{ostmoViewItRadiologist2023}. In this work, we extend the original study by introducing Window scaling and Random windowing, and by substantially expanding the analysis with additional experiments, ablations, metrics, and datasets.} \cite{ostmoViewItRadiologist2023}
adjusts which parts of the image distribution are visualized during training, and thus introduces useful variation into the training of DL models.

Window shifting stochastically adjusts the window level $L$ during preprocessing of training images, resulting in an augmentation effect after clipping. This is achieved by sampling a new window level, $L$ from a uniform distribution defined by $L_{\min}$ and $L_{\max}$
\begin{equation}
    L \sim \text{Uniform}(L_{\min}, L_{\max}).
    \label{eq:window_shifting}
\end{equation}

The boundaries of Window shifting, $L_{\min}$ and $L_{\max}$, can be set as hyperparameters or be determined from the distribution of foreground intensities in the CT dataset, tailored to the task at hand \cite{ostmoViewItRadiologist2023}.

\subsection{Window scaling}
Window shifting exploits the variation of HU shifts from contrast-enhancement in the dataset to augment the images. However, it does not account for uneven distribution of contrast agent within a foreground region, which may result in a tight or wide spread of HU for an image.

To account for this, and exploit the effect during training, we introduce \textit{Window scaling}. Window scaling scales the window width before clipping to vary how much of the image distribution is included during training, resulting in an augmentation effect. Specifically, the CT images are clipped with a randomly sampled width $W$, from a uniform distribution
\begin{equation}
    W \sim \text{Uniform}(W_{\min}, W_{\max}),
    \label{eq:window_scaling}
\end{equation}
where $W_{{\min}}$ and $W_{{\max}}$ are the minimum and maximum widths for the augmentation strength. We sample $W$ from a range around the base width. Hence, $W_{\min} \leq W_{\text{base}} \leq W_{\max}$. This allows the Window scaling to yield continuous variations around the base width. This makes it natural to use the base window during inference.

The resulting augmentation effect is, in some settings, similar to standard intensity scaling and contrast enhancement. However, as the augmentation happens before clipping, similar to Window shifting, the output is not limited by the initial preprocessing setting, which may cause artifacts.

\subsection{Random windowing}
Window shifting and Window scaling both work on independent parameters of the viewing window, allowing them to be combined without overhead.
We refer to the combined transformation of Window shifting and scaling as \textit{Random windowing}, due to the randomness introduced in the selection of both window level and width. The computational cost is negligible as it is performed in place of standard windowing.
Following common augmentation practices, we sample $L$ and $W$ independently, with probability $p_L$ and $p_W$, from uniform distributions, but acknowledge the potential for more data driven approaches. Our preliminary exploration in this direction did not lead to significant improvements, but we encourage further investigation in future work.

We present the combined preprocessing and augmentation technique of Random windowing, using both Window shifting and Window scaling, in Algorithm \autoref{alg:random_windowing}\footnote{Code at \url{https://github.com/agnalt/random-windowing}.}.

\begin{algorithm} 
    \caption{Random windowing algorithm} 
    \begin{algorithmic} 
    \State $x \gets ct\_image$ \Comment{In Hounsfield units} 
        \State $W \gets base\_width$ 
        \State $L \gets base\_level$ 
        \If{$\text{uniform}(0, 1) < p_W$} 
            \State $W \gets \text{uniform}(W\_min, W\_max)$ \Comment{Window scaling} 
        \EndIf 
        \If{$\text{uniform}(0, 1) < p_L$} 
            \State $L \gets \text{uniform}(L\_min, L\_max)$ \Comment{Window shifting} 
        \EndIf 
        \State $lower \gets L - W/2$ 
        \State $upper \gets L + W/2$ 
        \State $x \gets \text{clip}(x, lower, upper)$  \Comment{Windowing}
        \State $x \gets (x - lower) / W $ \Comment{Normalize to zero-one} 
    \end{algorithmic} 
    \label{alg:random_windowing} 
\end{algorithm}

\section{Analysis of Random windowing}
\label{sec:analysis_of_random_windowing}
The following sections explore how Random windowing improves and intentionally distorts images, avoids augmentation artifacts, and creates realistic yet challenging training samples. We also examine its impact on HU measurements and intensity distributions, highlighting its role in enhancing model performance and generalization.

\subsection{Image correction}
\label{sec:image_correction}
\begin{figure*}[t] % use figure* for spanning two columns
    \centering
    % First subfigure
    \begin{subfigure}{\columnwidth}
        \centering
        \includegraphics[width=\linewidth]{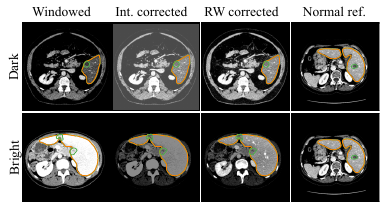}
        \caption{Improving visualization of difficult scans.}
        \label{fig:window_correction}
    \end{subfigure}
    \hfill
    % Second subfigure
    \begin{subfigure}{\columnwidth}
        \centering
        \includegraphics[width=\linewidth]{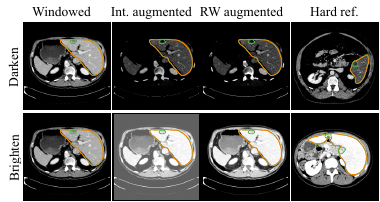}
        \caption{Simulating scans with non-standard contrast-enhancement.}
        \label{fig:window_distorted}
    \end{subfigure}
    \caption{Comparison of Random windowing and intensity augmentations. Random windowing samples beyond default window boundaries, improving visualizations during training, and recovering information lost with standard augmentations. It also produces realistic, challenging samples without the artifacts introduced by standard intensity transformations.}
    \label{fig:windowing_results}
\end{figure*}

Although CT scans are obtained with similar protocols, variations due to contrast-enhancement are expected. In \autoref{fig:window_correction}, "Windowed" and "Normal ref." display how the same clipping setting can result in different liver brightness in CT images due to contrast-enhancement. 

As Random windowing introduces variation to the CT clipping during training, it enables scans to be visualized in multiple ways, which can result in better visualizations. Intensity augmentations that transform clipped HU distributions will struggle to create the same variation.

In \autoref{fig:window_correction}, we aim to remedy the poorly timed contrast-enhancement using standard intensity augmentations and Random windowing. Standard augmentations cannot correct the loss of detail in the image, while the Random windowing settings yield a much better result. Additionally, standard intensity augmentations transform all values equally, and the background and bone structures, like the spine, outside the soft tissue range, are artificially darkened/brightened and can be considered artifacts in the final image.

\subsection{Image distortion}
An important task of data augmentation is to expose the model to images that resemble challenging training cases, so it can learn to generalize to difficult cases. Similar to how Random windowing can yield better visualizations of challenging images (Section \ref{sec:image_correction}), it can make normal training images look like the challenging ones, without introducing artifacts.

In \autoref{fig:window_distorted}, a CT slice where the liver has a normal response to contrast-enhancement is augmented to produce a training sample that resembles dark and bright training cases from the dataset. Standard intensity augmentations may fail to make realistic augmented images as they are prone to introducing artifacts in the background and bone structures. 

\subsection{Avoiding artifacts}
Artifacts from intensity augmentations in CT images occur when the pixel distribution is transformed after clipping. Particularly prone to causing such artifacts are intensity augmentations such as contrast augmentation, intensity scaling (i.e., brightness), and intensity shifting (i.e., additive brightness).

Artifacts occur when the edges of the intensity distribution are transformed such that they end up inside the original interval of $x$ (\autoref{eq:windowing}). In other words, the transformation $t$ moves $x_{\min}$ or $x_{\max}$ so
\begin{equation}
t(x_{\min}) > x_{\min} \quad \text{or} \quad t(x_{\max}) < x_{\max}.
\label{eq:artifacts}
\end{equation}

 As Random windowing performs augmentation through the window operation itself, it solves the problem of artifacts in \autoref{eq:artifacts}.

\subsection{Effect on HU measurements and intensity distribution}
\label{sec:distribution_and_context}
Until this point, the effect of Random windowing is mainly considered from an image perspective, where the pixel intensities are visualized as viewed by an observer. However, DL models process pixel values of the input and can, in principle, get strong clues from specific values. In the following paragraphs, we analyze the effect of Random windowing on the HU measurements and distribution of a CT scan. 

\subsubsection{Adjusted Hounsfield units}
For the CT modality, a unified global preprocessing scheme is beneficial during training to preserve information in the HU pixel measurements \cite{isenseeNnUNetSelfconfiguringMethod2021}. However, during augmentation, the HU are deliberately distorted to simulate useful variation and prevent overfitting.

Standard intensity augmentations do this by default on the input, while Random windowing obtains a similar effect through min-max normalization after clipping. Doing this resets the intensities to the zero-one range, ensuring that the HU are stochastically adjusted by the randomly sampled window width and level. 

In Section \ref{sec:additional_context_and_hu_ablation}, we verify that this step is key when working with tumor segmentation in contrast-enhanced CT images. However, skipping this step will allow Random windowing to preserve the absolute HU measurement in the scan while augmenting the image through added or removed context of the pixel distribution. In applications for CT without IV contrast, this might be beneficial as the original HU is intact.

\subsubsection{Additional context and characteristic distribution}
\begin{figure*}[]
    \centering
    \includegraphics[width=\linewidth]{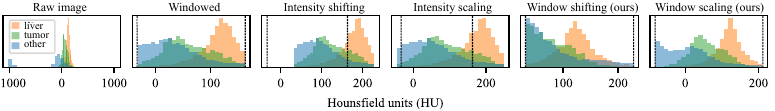}
    \caption{Augmentation effect on intensity distribution. Augmentation through intensity shifting and scaling affects the appearance of the image, but not the distribution shape. Shifting and scaling the viewing window can include more data near the edges of the base viewing window, so the shape of the distribution changes more.}
    \label{fig:shift_scale_hu_dist}
\end{figure*}

Regardless of whether HU are preserved or not, Random windowing can stochastically provide additional context compared to the clipped image view. Intensity augmentations are shown to be effective for certain DL applications as they prevent models from picking up on the characteristic distribution of the inputs \cite{chenSimpleFrameworkContrastive2020}. When linear augmentation transformations, like intensity shifting or scaling, are applied to the clipped intensity distribution, the absolute intensities are altered, but the relative shape of the distribution remains largely unchanged (\autoref{fig:shift_scale_hu_dist}).

Although Random windowing is parameterized by linear transformations in HU space, its effect on the final distribution can be non-linear. This is because the transformation of the window may expand the distribution by incorporating additional HU values, thereby reshaping the distribution rather than simply shifting or scaling it. This effect is further investigated in Section \ref{sec:additional_context_and_hu_ablation}. In the special case where Window scaling is performed with $W\sim \text{Uniform}(W_{\min}, W_{\text{base}})$ no additional context is included, and its effect is comparable to contrast augmentation with a scaling factor $\alpha \in (1, \frac{W_{\text{base}}}{W_{\min}})$ followed by clipping to the original range.

\section{Results}
\label{sec:results}
In this section, we empirically validate the effects of Random windowing in controlled experiments against traditional intensity-based augmentations from established baselines. Subsequently, we scrutinize the mechanisms at play in window augmentations and analyze the effect of base windows, augmentation components, and strengths.

\subsection{Stronger intensity augmentation pipeline}
We compare the proposed Random windowing augmentation against the intensity augmentation pipelines of two strong baselines, namely the nnU-Net \cite{isenseeNnUNetSelfconfiguringMethod2021} and the Unetr \cite{hatamizadehUNETRTransformers3D2022}. The intensity augmentations of the nnU-Net consist of contrast, multiplicative brightness, gamma, and inverse gamma augmentations applied in sequence on clipped and centered intensities. The Unetr applies intensity shifting and scaling of the clipped and zero-one-normalized intensities. We apply Random windowing with Window shifting and scaling independently on the raw CT intensities. In subsequent experiments, we standardize augmentation probabilities and strengths, but resort to recommended settings for each baseline here. Details in Appendix \ref{app:experiment_details}.

Each augmentation pipeline is used for training identical 3D-U-net \cite{cicek3DUNetLearning2016} segmentation models to perform liver tumor segmentation with 4-fold cross-validation on the Liver tumor segmentation (LiTS) dataset \cite{bilicLiverTumorSegmentation2023}. For robust evaluation, we consider the entire HepaticVessel (HV) dataset \cite{antonelliMedicalSegmentationDecathlon2022} (303 cases), Colorectal Liver Metastases (CRLM) dataset \cite{simpsonPreoperativeCTSurvival2024} (197 cases), and HCC-TACE dataset \cite{moawadMultimodalityAnnotatedHepatocellular2023} (104 cases) as disjoint test sets for liver tumor segmentation. With regards to tumor characteristics, HV and CRLM are more similar to the LiTS traning set than HCC-TACE. HCC-TACE comprises only patients with Hepatocellular carcinoma (HCC), where tumors show heterogeneous appearance due to variable tumor attenuation and portal venous washout. Due to the limited support in LiTS for HCC, HCC-TACE is especially difficult and in some degree out of domain.

For each prediction, we report the Dice similarity coefficient (DSC) measured with the original tumor mask, and report the mean performance in \autoref{tab:combined_augs} with the top performing method highlighted in bold. We measure the significance of the results with the Wilcoxon signed rank test at $p<0.05$. The results show that Random windowing leads to a statistically significant higher performance across all datasets.

\begin{table*}[t]
\centering
\caption{The mean DSC of the HV, CRLM, and HCC-TACE test sets. Random windowing significantly outperforms the intensity augmentation pipelines of the nnU-Net and Unetr. These results are consistent across whole datasets, as well as the difficult cases with low liver-tumor HU contrast and poor CE timing. ($^*$) denotes significance at $p < 0.05$.}
\label{tab:combined_augs}
\resizebox{\textwidth}{!}{%

\begin{tabular}{@{}r|lll|lll|lll@{}}
\toprule
                   & \multicolumn{3}{c|}{HepaticVessel}            & \multicolumn{3}{c|}{CRLM}                     & \multicolumn{3}{c}{HCC-TACE}                  \\
Intensity augmentation &
  \multicolumn{1}{c}{All} &
  \multicolumn{1}{c}{HU contrast} &
  \multicolumn{1}{c|}{CE timing} &
  \multicolumn{1}{c}{All} &
  \multicolumn{1}{c}{HU contrast} &
  \multicolumn{1}{c|}{CE timing} &
  \multicolumn{1}{c}{All} &
  \multicolumn{1}{c}{HU contrast} &
  \multicolumn{1}{c}{CE timing} \\ \midrule
None               & 0.507 ± 0.019 & 0.419 ± 0.027 & 0.365 ± 0.033 & 0.600 ± 0.006 & 0.449 ± 0.008 & 0.501 ± 0.006 & 0.305 ± 0.027 & 0.255 ± 0.023 & 0.144 ± 0.043 \\
Unetr (baseline)   & 0.527 ± 0.009 & 0.451 ± 0.010 & 0.395 ± 0.024 & 0.588 ± 0.021 & 0.438 ± 0.006 & 0.496 ± 0.031 & 0.329 ± 0.059 & 0.280 ± 0.060 & 0.196 ± 0.086 \\
nnU-Net (baseline) & 0.544 ± 0.026 & 0.476 ± 0.039 & 0.431 ± 0.028 & 0.606 ± 0.007 & 0.448 ± 0.014 & 0.528 ± 0.014 & 0.373 ± 0.070 & 0.313 ± 0.086 & 0.303 ± 0.071 \\
Random windowing (ours) &
  \textbf{0.566 ± 0.015$^*$} &
  \textbf{0.499 ± 0.017$^*$} &
  \textbf{0.450 ± 0.035$^*$} &
  \textbf{0.617 ± 0.003$^*$} &
  \textbf{0.471 ± 0.005$^*$} &
  \textbf{0.546 ± 0.023$^*$} &
  \textbf{0.393 ± 0.049$^*$} &
  \textbf{0.338 ± 0.054$^*$} &
  \textbf{0.333 ± 0.046$^*$} \\ \bottomrule
\end{tabular}%

}
\end{table*}

\subsection{Generalization to difficult tumor cases}
For an extended analysis of the augmentation pipeline results, we also measure the performance on what are considered difficult cases. The difficult cases are identified by \cite{bilicLiverTumorSegmentation2023}
% \citet
as images with low contrast between tumor and liver regions with mean tissue difference $< 20$ HU (HU contrast), in total 171, 42, and 68 cases for HV, CRLM, and HCC-TACE, respectively. Additionally, we identify that scans where the contrast-enhancement is poorly timed are difficult. Poor IV contrast timing can be identified by particularly high or low HU in the liver. By visual inspection, we consider the top and bottom 10 \% of scans with the highest and lowest median liver HU to be difficult, corresponding to HU $< 89$ and HU $> 137$, respectively (CE timing), in total 64, 39, and 16 cases for HV, CRLM, and HCC-TACE, respectively. In \autoref{tab:combined_augs} we report the mean DSC on these cases specifically, and find that models trained with Random windowing perform significantly better also on these subsets ($p < 0.05$).

\begin{figure}
    \centering
    \includegraphics{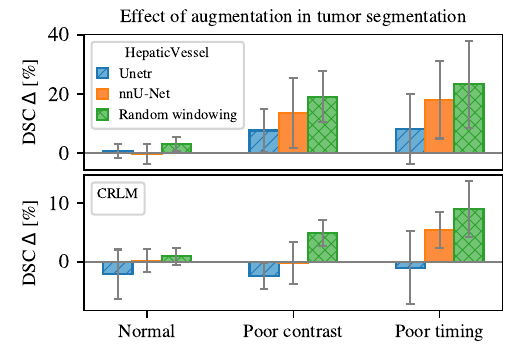}
    \caption{Relative DSC improvement by augmentation schemes measured for scans with normal contrast-enhancement, poor liver-tumor contrast, and poor contrast timing. The improvement is over not applying any intensity augmentations measured on the HepaticVessel and CRLM dataset.}
    \label{fig:dsc_improvement}
\end{figure}

To highlight the benefit of augmentation, we plot the relative improvement of DSC compared to not applying any intensity augmentations for the HV and CRLM datasets in \autoref{fig:dsc_improvement}. For comparison, we also compare with the "normal" scan subset, consisting of scans not part of the poor contrast or poor timing subsets. From evaluating \autoref{tab:combined_augs} and the relative improvements in \autoref{fig:dsc_improvement}, it is clear that Random windowing in general is helpful, but that conventional methods hurt performance in certain settings.
Compared to the baselines, Random windowing gives a larger improvement across all settings and is especially beneficial for difficult tumor cases, where the HU contrast is low or the timing is off. For HCC-TACE, we observe that augmentation and Random windowing are key due to the very limited support for HCC in the training set. Interestingly, Random windowing also benefits the normal cases across all datasets, more than the baseline alternatives. We hypothesize that this is due to its potential to use difficult cases to simulate normal cases as described in Section \ref{sec:image_correction}.

\subsection{Augmentation through context and HU adjustment}
\label{sec:additional_context_and_hu_ablation}
Compared to augmentation on clipped intensities, window augmentations can produce training samples with additional \textit{context} from the raw data. By context, we specifically refer to the parts of the CT intensity distribution that are near and outside the edges of the interval of the base window.

Although Random windowing does not preserve absolute HU by default, we hypothesize that context variation alone opens a new opportunity to augment CT intensities while preserving the HU of the image. We refer to this setting as Random windowing shift-scale (RW ss.), and is, to the best of our knowledge also novel and unexplored in CT augmentation.

To investigate this further, we ablate the effect of augmentation through additional context, as well as HU adjustments in Random windowing. HU adjustments are achieved through normalization (e.g., to $[0, 1]$) of the clipped and transformed intensities, and is common in standard intensity augmentations.

\begin{figure}
    \centering
    \includegraphics{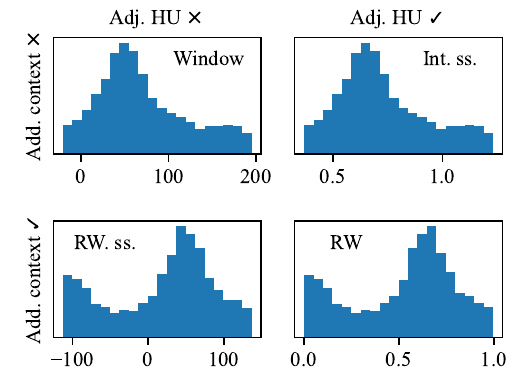}
    \caption{Illustration of the experiment settings used in the ablation of \autoref{tab:hu-context-ablation}. In each row, the overall shape of the distribution and the included HU values are the same. In each column, the HU are either preserved or not (scaled to $[0, 1]$).}
    \label{fig:hu-context-ablation}
\end{figure}

\autoref{fig:hu-context-ablation} illustrates the effects we are ablating with the distribution of one example scan. The initial row shows the distribution before and after augmentation when windowing is performed during preprocessing. In the second row, we augment the image while allowing additional context. 
For all settings, transformations are applied with $p=0.5$ and equal strengths on the z-score normalized to mean of 0 and standard deviation of 1 using the global dataset statistics.

\begin{table}[t]
\centering
\caption{Ablation of augmentation mechanisms in Random windowing. The experiment displays the additional benefit of adjusting Hounsfield units (Adj. HU) and providing additional data context (Add. cont.) during training augmentations. All other variables are unchanged. $^*$ indicates that the result is significantly larger than the next best alternative at $p<0.05$.}
\label{tab:hu-context-ablation}
\resizebox{\columnwidth}{!}{%

\begin{tabular}{@{}rccclccl@{}}
\toprule
\multicolumn{1}{l}{}      & Adj.     & Add.         & Aug-         &                     & \multicolumn{3}{c}{Instance-metrics}                         \\
                          & HU       & cont.      & mented       & Tumor DSC           & F1            & Recall        & Precision                    \\ \midrule
Base window    & $\times$ & $\times$     & $\times$     & 0.507 ± 0.019       & 0.592 ± 0.019 & 0.735 ± 0.032 & \textbf{0.624 ± 0.011$^{*}$} \\
RW shift-scale & $\times$ & $\checkmark$ & $\checkmark$ & 0.527 ± 0.008$^{*}$ & 0.582 ± 0.018 & 0.756 ± 0.011 & 0.586 ± 0.029                \\
Int. shift-scale &
  $\checkmark$ &
  $\times$ &
  $\checkmark$ &
  0.542 ± 0.024$^{*}$ &
  0.576 ± 0.025 &
  0.778 ± 0.024 &
  0.559 ± 0.031 \\
Random window &
  $\checkmark$ &
  $\checkmark$ &
  $\checkmark$ &
  \textbf{0.565 ± 0.017$^{*}$} &
  \textbf{0.604 ± 0.018} &
  \textbf{0.785 ± 0.019} &
  0.597 ± 0.034 \\ \bottomrule
\end{tabular}%
}

\end{table}

On the external test set, we measure the tumor DSC and the instance-wise lesion F1, recall, and precision, after a connected component analysis where $> 10 \%$ pixel overlap counts as a detected lesion. We present the results in \autoref{tab:hu-context-ablation}.

We observe that adjusting the HU has a larger impact than additional context, while both contribute constructively in Random windowing. We hypothesize that HU perturbations are important to guide the models away from HU reliance alone, as it increases tumor sensitivity. Meanwhile, augmentation in general decreases tumor precision, due to more false positives.

These results shed light on the mechanisms at play in Random windowing, while proving that the HU-preserving version of Random windowing is beneficial alone, and perhaps the only option in certain settings. We leave further exploration in this direction to future work.

\subsection{Importance of base viewing window}
\label{sec:effect_of_base_viewing_window}
A narrow viewing window enhances subtle intensity differences between liver tumors and surrounding parenchyma, but at the cost of reduced distribution context. The liver-tumor HU differences are emphasized by the HU shift of contrast-enhancement, which is exploited by Window shifting.

We hypothesize that using a region-specific narrow base window improves tumor segmentation by emphasizing the relevant HU differences. Furthermore, we expect Window shifting to benefit most when used with such focused windows.

To test this, we measure the impact of tumor and liver windows, covering 99 \% of foregrounds, as well as a window of raw HU and one characteristic of the general abdomen. We measure the impact of each window and its interaction with Window shifting in all settings.

\begin{table}[]
\centering
\caption{
Ablation study on the LiTS dataset reporting 2D validation tumor DSC (3 $\times$ repeated 4-fold CV). We observe that narrow, region-specific viewing windows improve tumor segmentation, and Window shifting further enhances performance, especially with focused windows.
}
\label{tab:window_ablation}
\resizebox{\columnwidth}{!}{%
\begin{tabular}{@{}rcccc@{}}
\toprule
Viewing window    & Width & Level & Baseline      & Window shifting \\ \midrule
None (raw)        & 2000           & 0              & 0.552 ± 0.081 & 0.580 ± 0.099   \\
Generic (abdomen) & 500            & 150            & 0.628 ± 0.078 & 0.636 ± 0.080   \\
Liver window      & 196            & 91             & 0.629 ± 0.091 & 0.637 ± 0.079   \\
Tumor window      & 169            & 65             & 0.634 ± 0.081 & 0.648 ± 0.084   \\ \bottomrule
\end{tabular}%
}
\end{table}

We report the window settings and tumor segmentation DSC in table \autoref{tab:window_ablation} and observe that both the baseline (static windowing) and Window shifting increase performance with narrower, more region-specific base windows. 
The performance gain is greatest when going from raw HU to a more focused window, even if only a generic soft tissue window. 

From \autoref{tab:window_ablation}, we observe that regardless of the base viewing window, Window shifting augmentation is advantageous. The results suggest that a sufficiently narrow window benefits Window shifting, and that the generic, liver, and tumor windows all are significantly better for Window shifting than the raw window, with $p<0.05$ using Wilcoxon's signed rank test between folds. 

\subsection{Robustness to augmentation strength}
\begin{figure}
    \centering
    \includegraphics[width=1\linewidth]{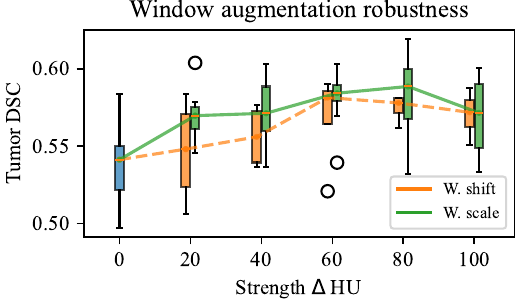}
    \caption{Window shifting and scaling improve tumor DSC at various strengths, with peaks at $L\pm 60$ and $W\pm 80$ HU.}
    \label{fig:param_robustness}
\end{figure}

\begin{figure}
    \centering
    \includegraphics[width=1\linewidth]{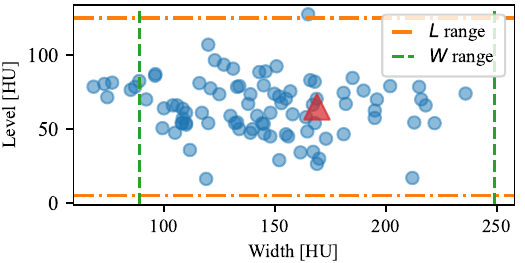}
    \caption{Per-case estimate of viewing windows, covering 99 \% of tumor HU, in the LiTS train set and base window ($\textcolor{red}{\blacktriangle}$). $\{L, W\}$ range show best shift/scale ranges from \autoref{fig:param_robustness}.}
    \label{fig:per_case_width_level}
\end{figure}

We measure the robustness to shifting and scaling parameters in Random windowing by independently varying the level shift and width expansion/reduction symmetrically around the base window setting. We perform 4 training runs for each strength and report the mean DSC on a 20 \% hold-out test set of the LiTS dataset. From \autoref{fig:param_robustness} we find both window shifting and scaling to improve performance at various strengths, with peak performance at window shift $\pm 80$ HU from the base window and width range $\pm 60$ HU. When plotting the per-case $(W,L)$ pairs we observe that these values correspond to natural variations in HU within the ROI for the dataset under consideration (\autoref{fig:per_case_width_level}).

\subsection{Effect of individual intensity augmentations}

\begin{table}[]
\centering
\caption{
Tumor DSC reported on the validation splits of the LiTS dataset and the independent HepaticVessel (HV) test set in 2D and 3D settings. Top performing methods are highlighted in bold.}
\label{tab:individual_augs}
\resizebox{\columnwidth}{!}{%
\begin{tabular}{@{}rcccc@{}}
\toprule
                      & LiTS tumor 2D & LiTS tumor 3D & HV tumor 2D & HV tumor 3D \\ \midrule
None (geometric only) & 0.634 ± 0.081   & 0.692 ± 0.087   & 0.445 ± 0.036 & 0.577 ± 0.041 \\
Intensity scaling     & 0.628 ± 0.091   & 0.650 ± 0.099   & 0.426 ± 0.046 & 0.516 ± 0.071 \\
Contrast              & 0.630 ± 0.088   & 0.668 ± 0.103   & 0.428 ± 0.039 & 0.553 ± 0.058 \\
Gamma                 & 0.635 ± 0.086   & 0.663 ± 0.132   & 0.480 ± 0.036 & 0.567 ± 0.058 \\
Gamma inverse         & 0.644 ± 0.083   & 0.669 ± 0.129   & 0.477 ± 0.035 & 0.568 ± 0.065 \\
Intensity shifting    & 0.632 ± 0.090   & 0.688 ± 0.104   & 0.455 ± 0.029 & 0.603 ± 0.040 \\
Window scaling (ours) &
  0.638 ± 0.091 &
  \bfseries 0.701 ± 0.077 &
  0.470 ± 0.025 &
  \bfseries 0.609 ± 0.037 \\
Window shifting (ours) &
  \bfseries 0.648 ± 0.084 &
  0.690 ± 0.089 &
  \bfseries 0.513 ± 0.031 &
  0.605 ± 0.041 \\ \bottomrule
\end{tabular}%
}
\end{table}

We experiment with two different DL architectures, the 3D-U-net \cite{cicek3DUNetLearning2016} and a slice-based 2D Deeplabv3+ \cite{chenEncoderDecoderAtrousSeparable2018}, to measure the robustness of individual transformation to various architectures. Each architecture is trained with geometric augmentations and one of the following intensity augmentations: 
contrast adjustment, intensity shifting (additive brightness), intensity scaling (multiplicative brightness), gamma adjustment, and gamma adjustment of inverse intensity values (gamma inverse). These augmentation methods are compared against the individual components of Random windowing augmentation, namely, Window shifting and Window scaling. All individual intensity augmentations are applied with the same probability.

The mean liver tumor DSC and standard deviations of 3 times repeated 4-fold cross validation are reported in \autoref{tab:individual_augs}.
The results show that the individual components of our method are indeed potent and surpass their intensity-based counterparts. Interestingly, applying no intensity augmentations (geometric only) outperforms individual intensity-based CT augmentations in certain settings, suggesting that some intensity augmentations may hurt performance.

\section{Limitations and future work}
\label{sec:discussion}
% Performance on non-contrast images
Window augmentations exploit the characteristics of contrast-enhanced CT scans, and throughout this study, we have focused on such images. While our technique often yields good results also on non-contrast-enhanced and low-quality CT images, the performance on such images is underexplored, and we have observed subpar performance. Segmenting images with no or low quality contrast-enhancement is very challenging due to the reduced contrast between tissue in such images, and due to the limited support in available labeled training data. Future work should explore and evaluate performance on these images, as they sometimes occur in clinical practice due to poor imaging or patient concerns.

% Potential beyond CT
Although Random windowing is tailored for the CT modality, its potential in other modalities is largely unexplored. Other imaging techniques with quantitative mappings of pixel values, such as Positron-Emission Tomography (PET) and standardized Magnetic Resonance Imaging (MRI) images, might benefit from training with Random windowing. These modalities have different physical properties and intensity distributions than CT, which may require further adaptation of the method. We leave this for future work.

% Road to clinical application
The potential in Random windowing underscores the importance of domain-specific augmentation techniques in medical imaging and the possibilities in clinical applications. Unlike generic augmentations adapted from natural image processing, CT-specific methods like Random windowing respect and build upon the unique properties of the modality, leading to more robust and performant models. As medical imaging remains a data-limited domain, robust augmentation techniques that meet the standards of the medical practice are crucial for DL and artificial intelligence to advance into the clinic.

\subsection{Conclusion}
In this study, we introduced Random windowing, a novel augmentation technique for CT images, and demonstrated its effectiveness in improving liver tumor segmentation performance. Random windowing enhances robustness in challenging tumor cases, particularly in scans with poorly timed contrast or low tumor-to-liver contrast.

Our results show that Random windowing consistently outperforms traditional intensity-based augmentations, such as those used in nnU-Net and Unetr, across multiple datasets, architectures, and metrics.
We attribute its generalization capabilities to the additional contextual information preserved from raw CT data, combined with HU adjustments that simulate natural variations in contrast-enhancement, allowing our method to utilize limited data efficiently.

Overall, Random windowing emerges as a powerful augmentation strategy for CT images, offering significant gains in segmentation performance under difficult imaging conditions. Future work could explore its extension to new applications, organs, and modalities, as well as its potential role in improving model robustness in clinical scenarios.

\bibliographystyle{IEEEtran}
\bibliography{main}
% \printbibliography

% \section*{Supplementary Material}

\appendix
\subsection{Experiment details}
\label{app:experiment_details}
All experiments are performed with the tumor base windows (\autoref{tab:window_ablation}) and augmentation strengths corresponding to $[L_{\min}, L_{\max}] = [12, 130]$ and $[W_{\min}, W_{\max}] = [129, 298]$, with total $p=0.3$ unless otherwise stated. All models are trained on the LUMI supercomputer, where 3D models operate on scans resampled to (1.5 mm)$^3$ voxel spacing, training with batch size 8 for 500 steps $\times$ 50 epochs and a patch size $96^3$. All models use geometric augmentations (random foreground crop and flipping along all axes with $p=0.5$), Batch normalization \cite{ioffeBatchNormalizationAccelerating2015}, AdamW optimizer \cite{loshchilovDecoupledWeightDecay2019}, deep supervision \cite{leeDeeplySupervisedNets2015}, and LeakyReLU activation \cite{leakyReLUxuEmpiricalEvaluationRectified2015} and learning rate 0.001. All models use the combined Cross-entropy and Dice loss \cite{isenseeNnUNetSelfconfiguringMethod2021}, without any class reweighing. 2D models follow the training recipe of \cite{ostmoViewItRadiologist2023}. 

\end{document}